\documentclass[pdflatex,sn-mathphys]{sn-jnl}
\usepackage{caption}
\usepackage{lscape}
\jyear{2022}

\begin{document}

\title[ ]{Application of Transformers based methods in Electronic Medical Records: A Systematic Literature Review}

\author*[1]{\fnm{Vitor} \sur{A. Batista}}\email{vabatista@ufrj.br}

\author[1]{\fnm{Alexandre} \sur{Evsukoff}}\email{alexandre.evsukoff@coc.ufrj.br}

\affil*[1]{\orgdiv{PEC/Coppe}, \orgname{Federal University of Rio de Janeiro}, \orgaddress{POBox 68506 21941-972}, \city{Rio de Janeiro}, \state{RJ}, \country{Brazil}}

\abstract{
The combined growth of available data and their unstructured nature has received increased interest in natural language processing (NLP) techniques to make value of these data assets since this format is not suitable for statistical analysis. This work presents a systematic literature review of state-of-the-art advances using transformer-based methods on electronic medical records (EMRs) in different NLP tasks. To the best of our knowledge, this work is unique in providing a comprehensive review of research on transformer-based methods for NLP applied to the EMR field. In the initial query, 99 articles were selected from three public databases and filtered into 65 articles for detailed analysis. The papers were analyzed with respect to the business problem, NLP task, models and techniques, availability of datasets, reproducibility of modeling, language, and exchange format. The paper presents some limitations of current research and some recommendations for further research.
}


\maketitle

\section{Introduction}\label{sec_introduction}
Hospitals, medical centers and clinics worldwide collect and record health information from their patients. Surveys from the US and UK show that 80 to 90\% of health establishments use an information system to store collected data \cite{OfHealth2018, Charles2015} into electronic medical records (EMRs). Furthermore, a fraction of this data is unstructured, since using only structured fields to input patient information is unfeasible \cite{Goossen2011}. Despite this, Wuster et. al. \cite{wurster2022analyzation} conducted a review on effects of EMR adoption and verified a positive impact for documentation.

The combined growth of available data and their unstructured nature has received increased interest in natural language processing (NLP) techniques to make value of these data assets since this format is not suitable for statistical analysis.

In the 1950s, when Noam Chomsky \cite{chomsky57a} published his work on syntactical grammars, an entire field was opened to explore new algorithms in human language. Until the 1980s, NLP algorithms used symbolic or rule-based methods: documents are parsed and compared to grammars, lookup tables, ontology or regular expressions \footnote{\url{https://en.wikipedia.org/wiki/Regular_expression}}) to extract important content or match corresponding structured data.  

Starting in late 1980s, statistical NLP came into the scene, including methods based on traditional machine learning using linguistic corpus to automatically extract information. In the 1990s, the birth and growth of the World Wide Web provided a rich source of data to build NLP models. The first attempts to represent words as numeric vectors came into play: latent semantic analysis (LSA) \cite{landauer1997solution} was used to create vectors by factorizing a matrix built from a word--word co-occurrence into documents. However, because of the exponential increase in corpus size, such methods needed large amounts of computational resources for processing. 

In the 2010s, a new and efficient technique, Word2Vec \cite{word2vec}, was developed to represent words as dense vectors, --which boosted a family of models for NLP: recurrent neural networks (RNNs), specifically, long short-term memory (LSTM) networks \cite{hochreiter1997long}. Although RNNs performed well on several NLP tasks when using Word2Vec representations as input, they suffer in training time and inference, since their intrinsic recurrence model cannot take advantage of parallelism.

Since 2017, the work from Vaswani\cite{Vaswani2017} and Delvin\cite{Devlin2018} created a revolution in NLP, establishing a new state of the art for several tasks using an architecture called transformers, which creates a flexible dense vector representation of words depending of their context in text and can be computed in an efficient (parallel) way using modern Graphical Processing Units (GPUs).

The purpose of this systematic literature review (SLR) is to provide the state-of-the-art advances using NLP to extract information and features for machine learning models from text into EMRs. Likewise, we want to understand limitations of and gaps in this research. 

It is well known that transformer-based models are currently the state of the art in almost all NLP tasks, from text classification to question answering. Since these models were first published in 2018 \cite{Devlin2018}, previous SLRs on NLP and EMRs could not provide a meaningful collection and analysis of work in the field. 

El Kah and Zeroual \cite{Zeroual2021} published a recent review on NLP applied to EHRs, but they did not conduct their review in a systematic way. Additionally, they chose the NLP tasks they wanted to survey and only a few selected medical problems. They also included studies using traditional machine learning techniques that have lower performance than transformers for NLP.

Unlike El Kah and Zeroual, Houssein et al. \cite{Houssein2021} conducted a systematic review of the field, but although they published their results in 2021, they did not included any work using transformers or BERT models. Also, Pandey and Janghel \cite{pandey2019recent} made a review of Deep Learning techniques on Health Systems, but didn't included transformers models into their analysis.

We also found other published reviews with too narrow of a scope on specific medical/business problems \cite{Tsang2020} \cite{Aryal2020} \cite{chadaga2022clinical} or on specific technology like big data \cite{alonso2017systematic}. 

Even though Laparra et. al. \cite{Laparra2021} conducted a great systematic review on NLP for EMRs, their work focused on identifying transfer learning techniques and also lacked important information such as the datasets used and their sizes, the business problems solved and the NLP tasks executed. 

To the best of our knowledge, our work is unique in that it provides a comprehensive review of state-of-the-art research on NLP applied to the EMR field. 

The remainder of this paper is organized as follows. Section \ref{sec_background} presents background on NLP and transformers (the focus of our research). In section \ref{sec_methods} we detail our research questions, data sources and methods we used to conduct this SLR. Section \ref{sec_results} summarize our data analysis. In section \ref{sec_discussion} we discuss the results, present some recommendations for future work as well point limitations of this study. Finally, in section \ref{sec_conclusion} we present our conclusion.

\section{Background}\label{sec_background}

\subsection{NLP}\label{subsec_nlp}
In the 1950s, NLP emerged as an interdisciplinary field covering computer science, artificial intelligence and linguistics. It is concerned with processing and analyzing human language to "understand" the contents of documents. One of the first applications of NLP was the IBM machine translation used to automatically translate Russian to English during World War II \footnote{\url{https://www.ibm.com/ibm/history/exhibits/701/701_translator.html}}. 

Traditionally, NLP is composed of stages: lexical analysis (or tokenization), syntactical analysis, semantics and pragmatics \cite{indurkhya2010handbook}. The lexical analysis step divides the text into tokens (i.e., words). This is particularly difficult for some languages, such as Chinese and Japanese. Syntax analysis gives meaning or the role of each token as sentences, providing a more suitable input for semantic analysis (literal meaning). Pragmatics is concerned with the disclosure of texts. This pipeline is typically called natural language understanding (NLU). Lexical and syntactical analyses are well established for many languages, but semantics and pragmatics still have room for improvement. In addition to NLU, which tries to obtain meaning and reason from natural language, there is natural language generation (NLG), comprising other groups of tasks, whereas algorithms generate new text. 

NLP comprises several different tasks. Some of them concern either NLU or NLG, such as conversational agents (i.e., chatbots). We can highlight some of these tasks as being important for the health domain:
\begin{itemize}
    \item Text Classification: Which label (or labels) best matches a given text, paragraph or document is predicted. This task is used to identify diseases from patient records.
    \item Semantic Similarity: Given a pair of text, how similar they are is determined. 
    \item Named Entity Recognition (NER): This task 
categorizes 
select specific spans into text that identifies entities such as diseases, exams, symptoms, and dates.
    \item Named Entity Linking: This task links a given recognized entity to a knowledge base. For instance, if an NER task identifies a disease, this task maps it to the ICD-10 database.
    \item Relation Extraction: The relation between the identified entities is determined. An example relation is the cause--effect relation: a disease causes a symptom. 
    \item Question Answering: From a collection of documents or records and a query in the natural language from a user, the system provide answers obtained from documents.
    \item Information Retrieval: This task is similar to question answering, but instead of providing direct answers to questions, it ranks a list of relevant documents that may contain information related to the user's query.
    \item De-identification: This task aims to convert personal information from patients records, such as names, birth dates, and IDs, into anonymous data, preserving privacy.
\end{itemize}

\subsection{Transformers}\label{subsec_transformers}

A transformer is an model architecture based on the self-attention mechanism. Attention was  introduced by Bahdanau et al. \cite{bahdanau2014neural} for machine translation tasks, where new representations for token (word) representations are computed based on context (surrounding words). 

\begin{figure}
  \centering
  \includegraphics[width=0.8\textwidth]{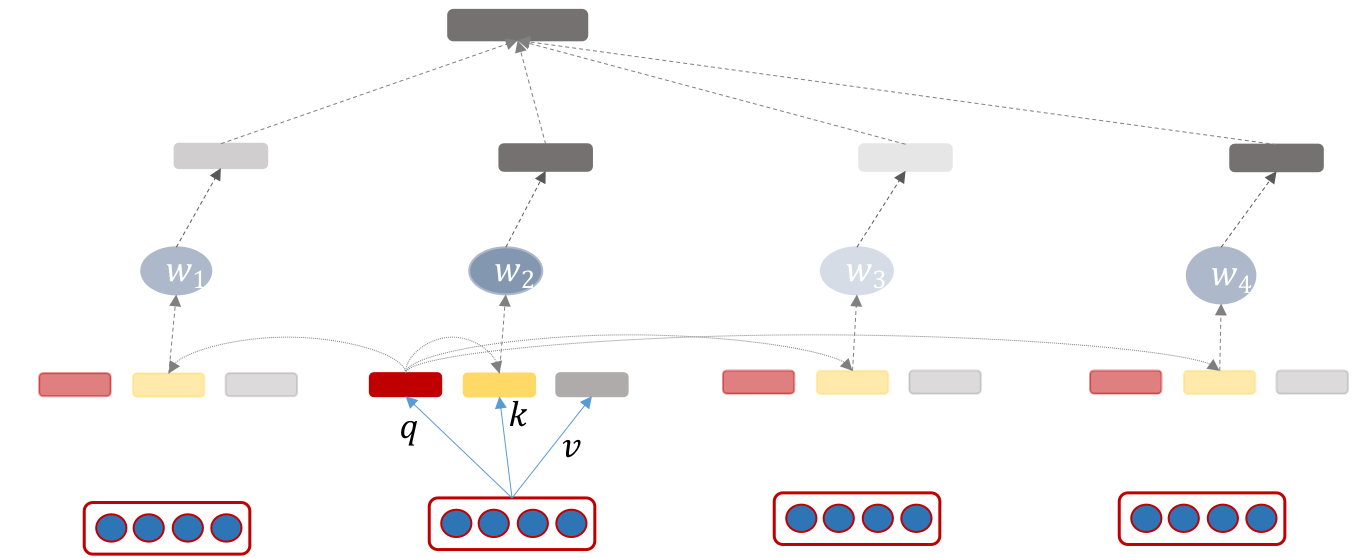}
  \caption{A head of self-attention mechanism}
  \label{figure:attention}
\end{figure}

A transformer block has several attention heads. Given a sequence of vector representations of input tokens represented by $x_{i}$ (Word2Vec representation), each attention head is computed as follows:
\begin{equation}
    q_{i}=W^Q.x_{i}, k_{i}=W^K.x_{i}, v_{i}=W^V.x_{i}
\end{equation}
\begin{equation}
    w_{i} = \frac{e^{q_{i}.k_{i}}}{\sum_{j=1}^{N} e^{{q_{i}.k_{j}}}}
\end{equation}
\begin{equation}
    Head_{n} = \sum w_{i}.v_{i}
\end{equation}

where $W^Q$, $W^K$ and $W^V$ are model parameters used to compute the \textit{query}, \textit{key} and \textit{value}, respectively, for each input $x_{i}$. $i$ represents the current token, and $j$ represents the token being attended to. $N$ is the input length. Figure \ref{figure:attention} illustrates this computation.

A transformer block aggregates several of these self-attention heads and feeds the aggregate forward into a fully connected layer. 

BERT (bidirectional encoder representations from transformers) uses the transformer architecture to pretrain a large language model, which is the result of an NLP task called language modeling, where the next words are predicted given a previous sequence. This is popular with autocomplete features in several software. The language model is trained using a very large corpus in a self-supervised way, without the expensive work of labeling data: next words are already in the text. A large corpus is easy to obtain: text can be obtained from Wikipedia, Wikibooks, the news and the entire Internet. Using transfer learning, BERT can learn specific NLP tasks (see the previous section) with few labeled samples (\textit{few-shot learning}). 

\section{Material and Methods}\label{sec_methods}

\begin{figure}
  \includegraphics[width=1.0\textwidth]{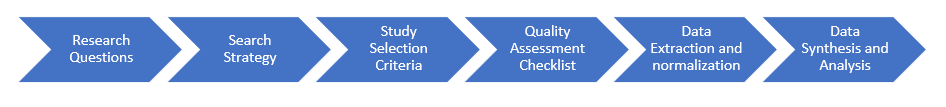}
  \caption{Workflow of our review process}
  \label{figure:workflow}
\end{figure}

We conducted this review using Kitchenham\cite{Kitchenham2004} guidelines for systematic literature reviews. Figure \ref{figure:workflow} depicts our review workflow.  In the next sections, we present our research questions, article selection criteria, queried databases and final collection of articles. 

\subsection{Research Questions}\label{subsec_questions}

To guide our efforts in this review, we propose the research questions below. 
\begin{enumerate}
  \item \textbf{Which business/medical problems are solved using transformers?}: We want to identify what kind of problems researchers are dealing with. Examples of problems include (but are not limited) mortality prediction, readmission prediction, information retrieval, and information extraction.
  \item \textbf{What NLP tasks are performed?}: What type of NLP task, including named entity recognition, relation extraction, and text classification, is studied? Which of them are well studied, and which lack research?
  \item \textbf{What models and techniques are used?}: In addition to BERT-based models, which techniques, models, and frameworks are researchers using to extract information from text?
  \item \textbf{What are the datasets}: What datasets are used as a basis for models? How large are these datasets? Are they publicly available?
  \item \textbf{Are studies reproducible?}: Do authors provide source code and data for reproducibility?
  \item \textbf{In which languages were the studies conducted?}: What other languages besides English are models trained on?
  \item \textbf{Do authors consume or produce results using standard data formats (i.e., HL7 FHIR\footnote{\url{http://hl7.org/fhir/}})?}: Is research producing results in standard format to be consumed by real-world applications?
\end{enumerate}

\subsection{Search Strategy}\label{subsec_search_strategy}

Since the object of our study is transformer-based models with unstructured text in EMRs, we propose this query be performed with academic search engines. We tried to cover main transformer types and variations of EMRs: "electronic health records" and "personal health records". 

\begin{verbatim}
    
("Transformer" OR "BERT" OR "GPT" OR "XLNet" OR "RoBERTa" OR 
"ALBERT" OR "LongFormer") AND ("EMR" OR "EHR" OR "PHR" OR 
"Electronic Medical Records" OR "Electronic Health Records" OR 
"Personal Health Records")

\end{verbatim}

Gusenbauer and Haddaway\cite{Gusenbauer2020} suggested 14 search engines suitable for systematic literature reviews. From this list, we picked PubMed, which is a superset of other databases, and the ACM Digital Library. We also included IEEExplore (not in the authors' list), as it is a very important academic database for computer science. The query was issued in December 2021. We initially found 99 articles in these three databases (see Figure \ref{figure:article_selection}).

\subsection{Selection Criteria}\label{subsec_selection_criteria}

We defined the criteria below to include articles in our review process: 
\begin{itemize}
    \item Articles must be written in English.
    \item The study must be performed using transformers.
    \item Articles must be freely available or available through an academic license. 
    \item Articles must use unstructured text as one of the inputs for models. Studies that used only structured data were not included.
    \item Articles cannot be another literature review.
    \item Articles must be peer-reviewed.
\end{itemize}

\begin{figure}
  \includegraphics[width=\linewidth]{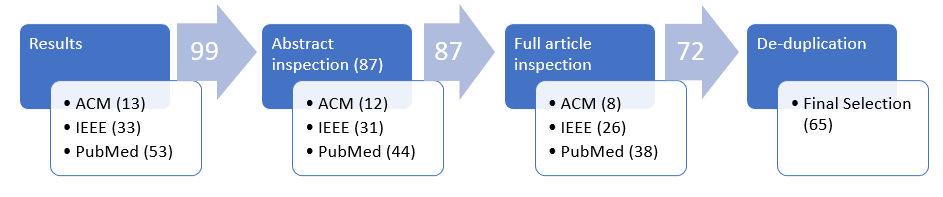}
  \caption{Flowchart of the selection process}
  \label{figure:article_selection}
\end{figure}

After collecting results, we read all abstracts to decide which articles were to be read and analyzed. At this point, we collected 12 articles from reviews. For instance, some of the articles were clearly studies on power transformers (energy domain). We did not eliminate any article if we were unsure whether it matched our criteria. After reading the remaining full articles, we eliminated 15 articles from our review because they either did not match the criteria or did not have enough information to answer our research question (i.e., short papers). The summary of this process is shown in Figure \ref{figure:article_selection}.

\subsection{Data extraction}\label{subsec_data_extraction}

After reading each paper, we carefully recorded information in a spreadsheet to answer our research questions. In addition, we reorganized each column individually (i.e., dataset name) into group information in a meaningful way. For example, the dataset size was grouped into a logarithm scale histogram for further analysis. 

\begin{figure}
  \centering
  \includegraphics[width=0.4\linewidth]{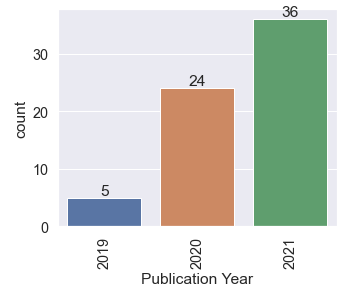}
  \caption{Publication year distribution}
  \label{figure:year_hist}
\end{figure}

The publication year of the 65 articles included for detailed reading is shown in Figure \ref{figure:year_hist}. As expected, the increasing number of publications shows an increase in interest in the field.

\section{Results}\label{sec_results}

In this section, we describe data collected from papers and analyze them using the research questions proposed in section \ref{subsec_questions}. Some of histogram plots do not add up to 65, i.e., the number of papers. This is because some papers produce more than one piece of information. For instance, some papers deal with more than one NLP task or business problem.
For transparency, we published all articles, references, spreadsheets and analysis at our article's GitHub repository: \url{https://github.com/vabatista/slr_transformers_ehr}.

\subsection{Business problem}\label{subsec_business_problem}

\begin{figure}
  \includegraphics[width=\linewidth]{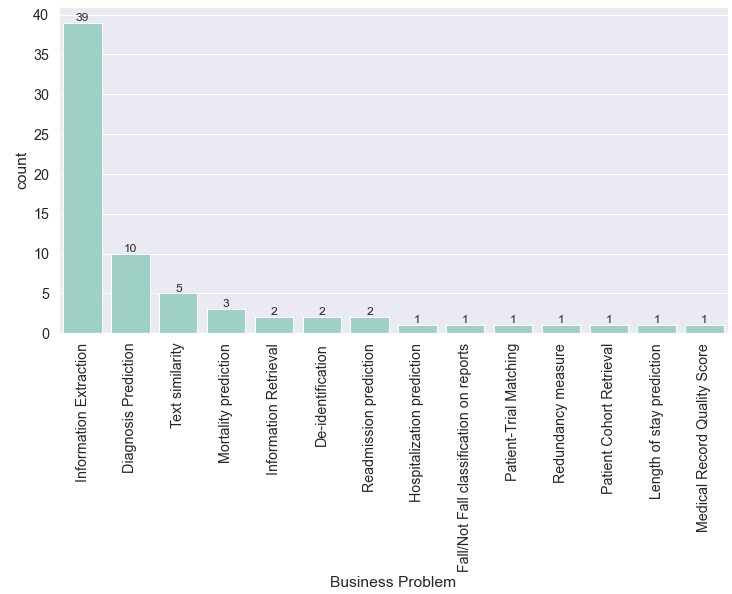}
  \caption{Business problem distribution}
  \label{figure:business_problem_hist}
\end{figure}

We define a business problem as the medical problem researchers are trying to solve. A business problem is solved by one or more NLP tasks. One example is hospital readmission prediction (if a patient will return to the hospital in the next X days), which can be solved by text classification task. Regarding business problems authors are dealing with, we can observe a concentration of studies in information extraction from unstructured information in patient text records (55\%). This is an important problem to create statistics and structured information from patient text records. In the second place, we can group predictions problems (10 diagnosis problems, 3 mortality problems, 2 readmission problems, and 1 hospitalization problem), accounting for 23\% of studies. 
Figure \ref{figure:business_problem_hist} shows the complete distribution of business problem authors are dealing with.

\subsection{NLP Task}\label{subsec_nlp_task}

\begin{figure}
  \includegraphics[width=\linewidth]{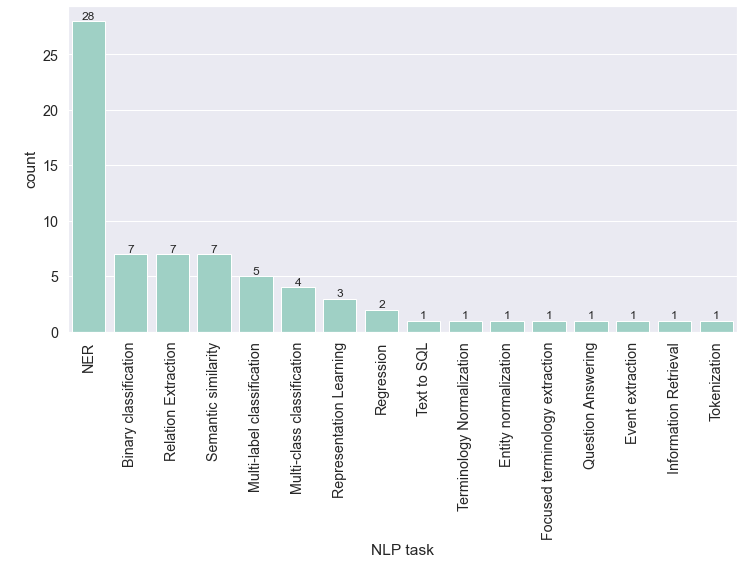}
  \caption{NLP task distribution}
  \label{figure:nlp_task_hist}
\end{figure}

Looking into the NLP task distribution (see Figure \ref{figure:nlp_task_hist}), the first two places are named entity recognition and relation extraction (49\% in total). This is consistent with the distribution of business problems. Classification problems (binary, multiclass and multilabel) represent 22.5\% of all studies. We can also see 7 studies dealing with semantic similarity (10\%). 
One study addresses the tokenization problem. Although this task is a preprocessing step for others, this is an important concern in the Chinese language.

\subsection{Models and Techniques}\label{subsec_models}

\begin{figure}
  \includegraphics[width=\linewidth]{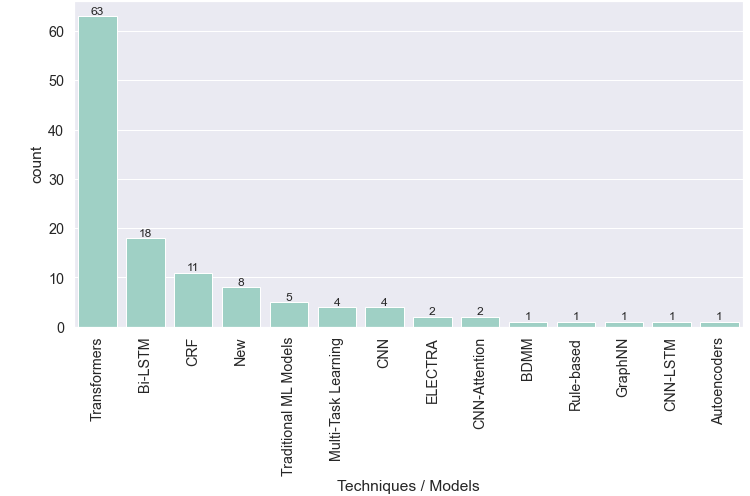}
  \caption{Techniques and model distribution}
  \label{figure:technique_distribution}
\end{figure}

To analyze models and techniques used or proposed in our literature review, we grouped them in a meaningful way. For instance, every type of transformer was grouped. In addition, we identified new models proposed as "New". The summary of data collected is shown in Figure \ref{figure:technique_distribution}. Several articles used more than one model or technique. Percentages were calculated using the total number of techniques found instead of the number of articles.

As expected, transformer-based models prevail over other types (51.6\%), whereas BERT and BioBERT are the most representative versions. Novel models represent only 6.5\% of all methods. For NER tasks, a combination of transformers, Bi-LSTM and CRF is present in 9 out of 28. Traditional ML models as accounted for 4\% of the models, used mostly for comparison. Convolutional neural networks and multitask learning account for 3.2\%. ELECTRA, which uses a generative adversarial strategy to pretrain a transformer, was mentioned in two articles. 

\subsection{Data and reproducibility}\label{subsec_datasets}

\begin{figure}
  \centering
  \includegraphics[width=0.5\textwidth]{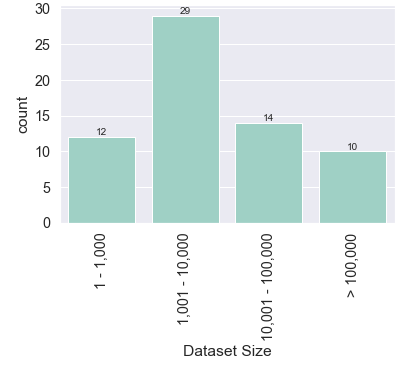}
  \caption{Dataset size range}
  \label{figure:dataset_size_hist}
\end{figure}

Regarding dataset size, it was difficult to compare datasets in several occasions. NER papers had sentence and entity counts and made comparisons to semantic similarity, in which a whole document or sentence is not symmetric. Furthermore, few studies used more than one dataset. In this case, we used the largest dataset as a measure. We made our best effort in normalizing sizes between articles and used the sentence as a base measure as much as possible. We also set the ranges of size into a logarithmic scale because of the variety of sizes we found. Figure \ref{figure:dataset_size_hist} shows the distribution across the ranges we defined. 

\begin{figure}
  \centering
  \includegraphics[width=0.5\textwidth]{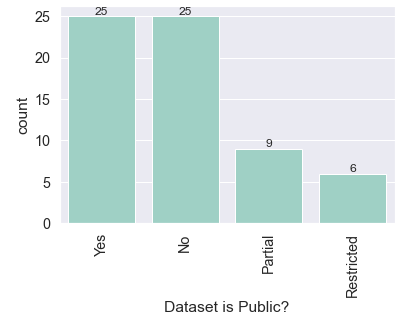}
  \caption{Dataset availability distribution}
  \label{figure:public_dataset_hist}
\end{figure}

\begin{figure}
  \centering
  \includegraphics[width=1\textwidth]{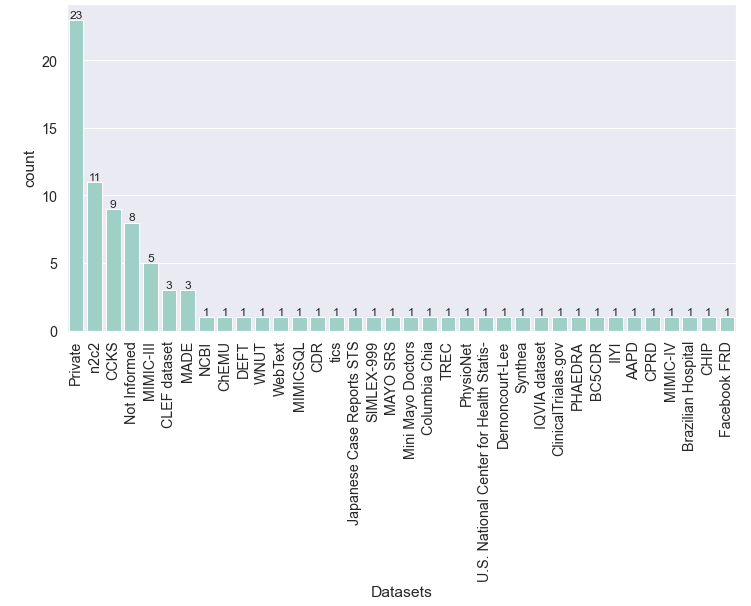}
  \caption{Datasets used}
  \label{figure:datasets_hist}
\end{figure}

We categorized dataset availability into public, private, partial and restricted. \textbf{Public} datasets are generally available to any researcher. Examples include but are not limited to China Conference on Knowledge Graph and Semantic Computing (CCKS) evaluation tasks, Harvard n2c2 \footnote{\url{https://portal.dbmi.hms.harvard.edu/projects/n2c2-nlp/}} (formerly i2b2) and  MIMIC-III \footnote{\url{https://archive.physionet.org/physiobank/database/mimic3cdb/}}. \textbf{Private} datasets are those not available to the research community. Most private datasets belong to private hospitals. Sometimes, not even the name of the hospital is mentioned in the study. \textbf{Partial} datasets are dataset where the authors used a mixture of public and private datasets. \textbf{Restricted} datasets are those that need approval and/or payment to obtain data (i.e., CPRD \footnote{\url{https://cprd.com/research-applications}} from the UK Public Health System). Additional information about dataset availability is discussed in the next section. Figure \ref{figure:public_dataset_hist} shows the distribution of papers using this criteria. Figure \ref{figure:datasets_hist} displays the datasets we found on published papers.

One of our research questions is about study reproducibility, which is as important as the results themselves, for peer review and validation. We evaluated when an author made the code available in a public repository (i.e., GitHub) and also the availability of the used datasets. 

\begin{table}[!ht]
    \centering
    \caption{Dataset vs code availability distribution}
    \begin{tabular}{cccccc}
    
        ~ & \multicolumn{4}{c}{Dataset availability} & ~ \\ \cline{2-6}
        Source code availability & Public & Private & Partial & Restricted & Total \\ \hline
        Not informed & 20\% & 35\% & 9\% & 6\% & 71\% \\
        Public & 18\% & 3\% & 5\% & 3\% & 29\% \\ \hline
        Total & 38\% & 38\% & 14\% & 9\% & ~ \\ 
    \end{tabular}
    
    \label{table:reproducibility}
\end{table}
Table \ref{table:reproducibility} summarizes our findings. Only 29\% of authors made their source code available. 18\% of studies are fully reproducible. A relevant portion of studies provided neither the source code nor datasets used (35\%).

\begin{figure}
  \centering
  \includegraphics[width=0.8\textwidth]{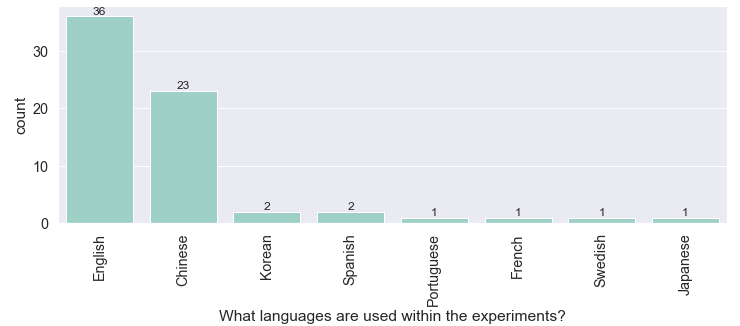}
  \caption{Languages vs availability of datasets}
  \label{figure:language_hist}
\end{figure}

The majority of studies were conducted in English (53\%) and Chinese (23\%). Other languages include Spanish, Korean, Portuguese, Swedish, French and Japanese. We discuss this phenomena in section \ref{sec_discussion}. Figure \ref{figure:language_hist} shows a histogram of the datasets of languages. 

Unfortunately, none of the studies provide information about the usage of standard formats, such as HL7 FHIR, for consumption or as an output format of their work. 

\section{Discussion}\label{sec_discussion}

Our SLR highlights the increasing interest in transformer-based models to extract unstructured information from EMRs. However, we also found several issues with studies that can help guide future research efforts. 

One finding that is a major concern is reproducibility. Only 18\% of studies are fully reproducible. This weakens peer review processes and the reliability of research. We know that EMRs from hospitals and clinics have sensitive data and cannot be made public. On the other hand, only 2 out of 65 studies are dedicated to sensitive data de-identification. 

\begin{figure}
  \centering
  \includegraphics[width=1\textwidth]{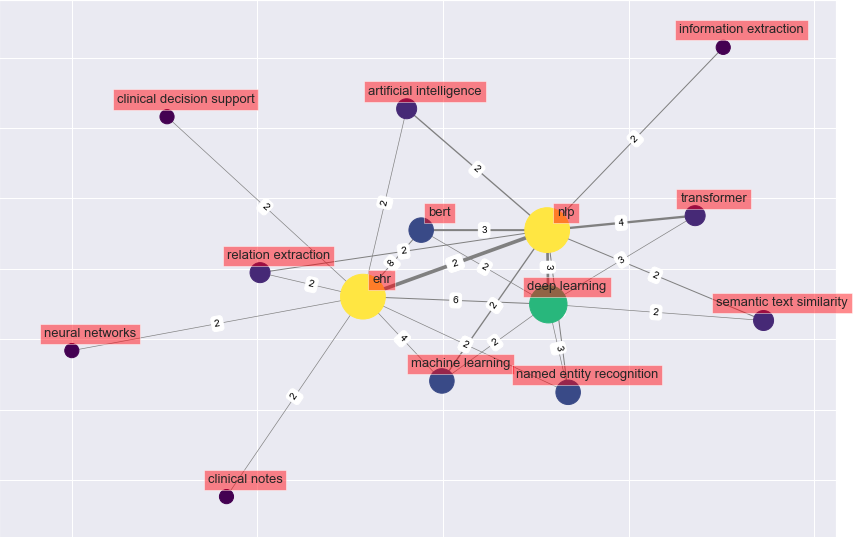}
  \caption{Keywords co-occurrence graph}
  \label{figure:word_graph}
\end{figure}

Additionally, we observed a concentration of efforts on the information extraction problem, more specifically focused on the named entity recognition task. Another aspect that deserves attention is that 88\% of studies are conducted in English or Chinese. An interesting finding was the presence of few novel models to solve problems. Most studies concern into hyperparameter tuning using established technologies over public benchmarks and datasets. Analyzing the co-occurrence  of words present in article keywords (Figure \ref{figure:word_graph})\footnote{Only keywords that co-occurred more than once.}, we also confirmed more interest in NER tasks than other NLP tasks. These aspects are related to benchmark dataset availability. CCKS, Harvard n2c2 and MIMIC-III (see section \ref{subsec_datasets}) are present in a majority of published works. 

As pointed out by Roberts et al. \cite{roberts2021common}, demographics of data are important to understand the robustness of model predictions. For instance, can a model trained with data from one hospital use data from another to test predictions? During our analysis, we identified only five articles that explicitly describe the demographics of the dataset used.

None of the studies declared using industry standard formats for consumption or output the formats of their models. This shows the gap between academic research and industry necessities, since research results need substantial software engineering for production. 

Based on the issues we encountered during our systematic review, we propose the following recommendations.
\begin{enumerate}
\item To move forward in the field, researchers must prioritize  de-identification tasks to turn private datasets into public datasets. As a result, more data will be available for labelling and supervised learning tasks. Moreover, dataset languages will become more diverse, and studies will be more reproducible. 
\item Codes should be public even if a dataset is not. Researchers could compare the results of models using their own private datasets to confirm or refute the original hypothesis. 
\item Models and datasets should be made available using industry standard formats such as HL7 FHIR. As a result, the software industry should take advantage of research results faster and invest more in new research.
\item Datasets should include demographic information from patients, even anonymized information. This will help researchers create more reliable models.
\end{enumerate}

\textbf{Limitations}. In this study we chose 3 important article databases to issue our query: IEEExplore, the ACM Digital Library and PubMed. We exposed the reasons in section \ref{subsec_search_strategy}. It is known that several papers are published at \url{arXiv.org}, including the original transformer and BERT articles. This could lead to important work not covered by our review. However, since databases such as arXiv are not peer-reviewed, the number of publications has grown exponentially, making any systematic literature review impractical. Similarly, in our query (detailed in section \ref{subsec_search_strategy}), we did not include terms for all types of BERT- or transformer-based models, because an extensive list is itself a research task. 

\section{Conclusion}\label{sec_conclusion}
In this SLR, we collected and summarized 65 research papers using transformer-based models to solve NLP tasks using EMRs. We analyzed the type of business problems and NLP task to be solved, the reproducibility, detailed information about the datasets used and the models applied. We made all collected data publicly available on GitHub for research community scrutinization.

In summary we found that research is driven by public dataset availability. For this reason, we observed a concentration of work on English and Chinese NER tasks. We also made recommendations to advance research in the field faster: focusing on de-identification technology to make more data available and providing industry standard formats for ease of adoption. 

\section{Declarations}\label{sec_declarations}

\subsection{Ethical Approval and Consent to participate}\label{subsec_ethical}
Not applicable.

\subsection{Human Ethics}\label{subsec_human_ethics}
Not applicable.

\subsection{Consent for publication}\label{subsec_consent}
Not applicable.

\subsection{Availability of supporting data}\label{subsec_availability_data}
All collected data and analysis are available through our Github repository: \url{https://github.com/vabatista/slr_transformers_ehr}

\subsection{Competing interests}\label{subsec_competing}
The authors declare that they have no conflict of interest.

\subsection{Funding}\label{subsec_funding}
This research had no funding.

\subsection{Authors' contributions}\label{subsec_authors}
The authors contributed equally to this work.

\subsection{Acknowledgements}\label{subsec_ack}
We would like to thanks to Federal University of Rio de Janeiro to provide subscriptions to article databases access.

\subsection{Authors' information}\label{subsec_authors_inf}
\textbf{Computer Scientist and PhD Candidate at Federal University of Rio de Janeiro, Rio de Janeiro, Brazil}.

\noindent Vitor A. Batista

\noindent
\textbf{Associate professor at Federal University of Rio de Janeiro, Rio de Janeiro, Brazil}.

\noindent Alexandre Evsukoff


\section{Appendix - Selected references for SLR}
\resizebox{\textwidth}{!}{
\begin{tabular}{lllllll}
\toprule
                   Ref. & Business problem & NLP Task & Dataset size & Dataset Public? & Code Public? &        Languages \\
\midrule

                \cite{Pan2021} &                              Information Retrieval &                                     Text to SQL &     1,001 - 10,000 &            Yes &         Yes &          English \\
\cite{10.1145/3436369.3436390} &                             Information Extraction &                                             NER &          1 - 1,000 &            Yes &          No &          Chinese \\
                \cite{9017143} &                             Information Extraction &                                             NER &     1,001 - 10,000 &            Yes &          No &          Chinese \\
              \cite{Liang2021} &                             Information Extraction &                       Terminology Normalization &     1,001 - 10,000 &            Yes &         Yes &          Chinese \\
              \cite{Finch2021} &   Hospitalization prediction, Mortality prediction &                           Binary classification &          > 100,000 &             No &          No &          English \\
                \cite{9474713} &            Fall/Not Fall classification on reports &                           Binary classification &     1,001 - 10,000 &            Yes &         Yes &       Portuguese \\
                \cite{Wan2020} &                             Information Extraction &                                             NER &     1,001 - 10,000 &        Partial &          No &          Chinese \\
      \cite{Christopoulou2020} &                             Information Extraction &                             Relation Extraction &   10,001 - 100,000 &            Yes &          No &          English \\
                 \cite{Li2021} &                              Information Retrieval &                             Semantic similarity &     1,001 - 10,000 &            Yes &          No &          English \\
                \cite{9565737} &                               Mortality prediction &                           Binary classification &     1,001 - 10,000 &            Yes &         Yes &          English \\
                \cite{Luo2021} &                               Diagnosis Prediction &                           Binary classification &          > 100,000 &             No &          No &          English \\
               \cite{Shin2021} &                               Diagnosis Prediction &                      Multi-class classification &          > 100,000 &             No &          No &           Korean \\
                \cite{Li2020a} &                               Diagnosis Prediction &                      Multi-class classification &          > 100,000 &     Restricted &          No &          English \\
                \cite{9037721} &                               Diagnosis Prediction &                      Multi-label classification &   10,001 - 100,000 &        Partial &          No &          Chinese \\
               \cite{Meng2021} &                               Diagnosis Prediction &                           Binary classification &   10,001 - 100,000 &             No &         Yes &          English \\
                \cite{8965108} &                             Information Extraction &                                             NER &          1 - 1,000 &            Yes &          No &          English \\
              \cite{Mitra2020} &                             Information Extraction &                                             NER &     1,001 - 10,000 &             No &          No &          English \\
                 \cite{Li2020} &                             Information Extraction &                                             NER &   10,001 - 100,000 &            Yes &         Yes &          Chinese \\
                \cite{9602374} &                             Information Extraction &                                             NER &     1,001 - 10,000 &            Yes &          No &          Chinese \\
                 \cite{Wu2021} &                             Information Extraction &                                             NER &   10,001 - 100,000 &             No &          No &          Chinese \\
                \cite{9581252} &                                  De-identification &                                             NER &     1,001 - 10,000 &            Yes &          No &          English \\
               \cite{Gong2020} &                             Information Extraction &                                             NER &          1 - 1,000 &     Restricted &          No &          Chinese \\
                \cite{9361169} &                             Information Extraction &                        NER, Relation Extraction &     1,001 - 10,000 &             No &          No &          Chinese \\
                \cite{9648338} &                             Information Extraction &                             Relation Extraction &     1,001 - 10,000 &            Yes &          No &          English \\
                \cite{9313224} &                             Information Extraction &                           Binary classification &   10,001 - 100,000 &             No &          No &          English \\
\cite{10.1145/3366423.3380181} &                             Patient-Trial Matching &                         Representation Learning &          > 100,000 &        Partial &         Yes &          English \\
\cite{10.1145/3459930.3469547} &                             Readmission prediction &                         Representation Learning &   10,001 - 100,000 &     Restricted &         Yes &          English \\
            \cite{Johnson2020} &                                  De-identification &                                             NER &   10,001 - 100,000 &            Yes &         Yes &          English \\
                \cite{9628010} &                             Information Extraction &                     Multi-label classification  &          > 100,000 &            Yes &         Yes &          English \\
               \cite{Chen2021} &                               Diagnosis Prediction &                         Representation Learning &          > 100,000 &        Partial &          No &          Chinese \\
                \cite{9565714} &                               Diagnosis Prediction &                     Multi-label classification  &          > 100,000 &             No &          No &          Chinese \\
             \cite{Naderi2021} &                             Information Extraction &                                             NER &     1,001 - 10,000 &            Yes &          No &  English, French \\
             \cite{Searle2021} &                                 Redundancy measure &                             Semantic similarity &   10,001 - 100,000 &        Partial &         Yes &          English \\
              \cite{Jiang2020} &                               Diagnosis Prediction &                      Multi-class classification &     1,001 - 10,000 &             No &          No &          Chinese \\
                 \cite{Li2019} &                             Information Extraction &                            Entity normalization &          > 100,000 &            Yes &         Yes &          English \\
                \cite{9473700} &                             Information Extraction &                  Focused terminology extraction &   10,001 - 100,000 &             No &          No &          Swedish \\
                \cite{9565778} &                             Information Extraction &                                             NER &     1,001 - 10,000 &             No &          No &          English \\
            \cite{Mahajan2020} &                                    Text similarity &                             Semantic similarity &     1,001 - 10,000 &            Yes &          No &          English \\
                \cite{9565706} &                             Information Extraction &                                             NER &          1 - 1,000 &             No &          No &          English \\
\cite{10.1145/3410530.3414436} &                             Information Extraction &                              Question Answering &          1 - 1,000 &            Yes &          No &          English \\
                \cite{9529801} &                             Information Extraction &                                Event extraction &     1,001 - 10,000 &             No &          No &          Chinese \\
                \cite{Kim2020} &                             Information Extraction &                                             NER &          1 - 1,000 &             No &         Yes &           Korean \\
                \cite{9562815} &                             Information Extraction &                                             NER &     1,001 - 10,000 &            Yes &         Yes & English, Spanish \\
                \cite{9529670} &                             Information Extraction &                                             NER &     1,001 - 10,000 &             No &          No &          Chinese \\
                \cite{9123057} &                               Diagnosis Prediction &                      Multi-label classification &     1,001 - 10,000 &             No &          No &          Chinese \\
              \cite{Zhang2020} &                             Information Extraction &                                             NER &     1,001 - 10,000 &             No &          No &          Chinese \\
            \cite{Alimova2020} &                             Information Extraction &                             Relation Extraction &          1 - 1,000 &            Yes &         Yes &          English \\
\cite{10.1145/3374587.3374612} &                             Information Extraction &                                             NER &   10,001 - 100,000 &             No &          No &          Chinese \\
                \cite{9587454} &                             Information Extraction &                                             NER &     1,001 - 10,000 &             No &          No &          Chinese \\
                \cite{8965823} &                             Information Extraction &                                             NER &          1 - 1,000 &        Partial &          No &          Chinese \\
               \cite{Zhan2021} &                             Information Extraction &                        NER, Relation Extraction &          1 - 1,000 &            Yes &         Yes &          English \\
               \cite{Soni2020} &                           Patient Cohort Retrieval &                           Information Retrieval &   10,001 - 100,000 &     Restricted &          No &          English \\
           \cite{Khambete2021} &              Diagnosis Prediction, Text similarity & Multi-class classification, Semantic similarity &     1,001 - 10,000 &     Restricted &          No &          English \\
              \cite{Mitra2021} &                             Information Extraction &                             Relation Extraction &          1 - 1,000 &             No &          No &          English \\
\cite{10.1145/3404555.3404635} &                             Information Extraction &                             Relation Extraction &     1,001 - 10,000 &             No &          No &          Chinese \\
            \cite{Mutinda2021} &                                    Text similarity &                             Semantic similarity &     1,001 - 10,000 &        Partial &          No &         Japanese \\
             \cite{Darabi2020} & Length of stay prediction, Readmission predicti... &               Regression, Binary classification &   10,001 - 100,000 &     Restricted &         Yes &          English \\
          \cite{Steinkamp2020} &                             Information Extraction &                                             NER &     1,001 - 10,000 &             No &          No &          English \\
                \cite{Lin2021} &                       Medical Record Quality Score &                                      Regression &          > 100,000 &             No &          No &          English \\
               \cite{Nath2021} &            Information Extraction, Text similarity &                        NER, Semantic similarity &     1,001 - 10,000 &            Yes &          No &          English \\
\cite{10.1145/3459930.3469560} &                             Information Extraction &                                             NER &   10,001 - 100,000 &            Yes &         Yes &          English \\
                \cite{9430499} &                             Information Extraction &                Multi-label classification , NER &          1 - 1,000 &        Partial &         Yes &          Spanish \\
               \cite{Yuan2020} &                             Information Extraction &                                    Tokenization &          1 - 1,000 &             No &          No &          Chinese \\
              \cite{Xiong2020} &                                    Text similarity &                             Semantic similarity &     1,001 - 10,000 &            Yes &          No &          English \\
                \cite{9537857} &                             Information Extraction &                                             NER &     1,001 - 10,000 &        Partial &          No &          Chinese \\

\bottomrule
\end{tabular}

}
Full version available in: \url{https://github.com/vabatista/slr_transformers_ehr}.

\bibliography{sn-bibliography, sn-selected-references}

\end{document}